\documentclass[a4paper]{article}

\usepackage[margin=1in]{geometry} 

\usepackage{amsmath}
\usepackage{amsthm}
\usepackage{amssymb}

\usepackage[utf8]{inputenc}
\usepackage{hyperref}
\hypersetup{
	unicode,
	pdfauthor={Author One, Author Two, Author Three},
	pdftitle={A simple article template},
	pdfsubject={A simple article template},
	pdfkeywords={article, template, simple},
	pdfproducer={LaTeX},
	pdfcreator={pdflatex}
}

\usepackage[sort&compress,numbers,square]{natbib}
\theoremstyle{plain}

\theoremstyle{definition}

\usepackage{graphicx, color}
\graphicspath{{fig/}}

\usepackage{algorithm, algpseudocode} 
\usepackage{mathrsfs} 

\usepackage{lipsum}

\usepackage[disable]{todonotes}
\usepackage{adjustbox}
\usepackage{booktabs}
\usepackage{subcaption}
\usepackage{graphicx}
\usepackage{makecell}
\usepackage{placeins}

\title{Deep Survival Analysis of Longitudinal EHR Data for Joint Prediction of Hospitalization and Death in COPD Patients}
\author{Enrico Manzini$^{1,2,3}\thanks{Corresponding author: enrico.manzini@upc.edu} $
\and Thomas Gonzalez Saito$^{1,2,3}$ 
\and Joan Escudero$^4$,
\and Ana Génova$^4$,
\and Cristina Caso$^4$,
\and Tomas Perez-Porcuna$^{5,6}$,
\and Alexandre Perera-Lluna$^{1,2,3}$ 
}

\date{
	$^1$B2SLab, Institut de Recerca i Innovació en Salut (IRIS), Universitat Politècnica de Catalunya, Barcelona, Spain\\
	$^2$Networking Biomedical Research Centre in the subject area of Bioengineering, Biomaterials and Nanomedicine (CIBER-BBN), Madrid, Spain\\  
    $^3$Institut de Recerca Sant Joan de Déu, Esplugues de Llobregat, Barcelona, Spain\\
    $^4$Evidenze Health, Spain \\
    $^5$Servei de Vigilància Epidemiològica i Resposta a Emergències de Salut Pública a Barcelona Sud, Subdirecció Regional a Barcelona, Secretaria de Salut Pública, Departament de Salut, Generalitat de Catalunya, Spain\\
    $^6$Fundació Docència i Recerca Mútua Terrassa, Catalonia, Spain\\
}

\begin{document}

\maketitle

	\begin{abstract}
		Patients with chronic obstructive pulmonary disease (COPD) have an increased risk of  hospitalizations, strongly associated with decreased survival, yet predicting the timing of these events remains challenging and has received limited attention in the literature. In this study, we performed survival analysis to predict hospitalization and death in COPD patients using longitudinal electronic health records (EHRs), comparing  statistical models, machine learning (ML), and deep learning (DL) approaches. We analyzed data from more than 150k patients from the SIDIAP database in Catalonia, Spain, from 2013 to 2017, modeling hospitalization as a first event and death as a semi-competing terminal event. Multiple models were evaluated, including Cox proportional hazards, SurvivalBoost, DeepPseudo, SurvTRACE, Dynamic Deep-Hit, and Deep Recurrent Survival Machine. Results showed that DL models utilizing recurrent architectures outperformed both ML and linear approaches in concordance and time-dependent AUC, especially for hospitalization, which proved to be the harder event to predict.  This study is, to our knowledge, the first to apply deep survival analysis on longitudinal EHR data to jointly predict multiple time-to-event outcomes in COPD patients, highlighting the potential of DL approaches to capture temporal patterns and improve risk stratification.

        \noindent\textbf{Keywords:} COPD, hospitalization, mortality, survival analysis, deep learning, EHR
	\end{abstract}


\section{Introduction}\label{sec:intro}
\paragraph{}Chronic Obstructive Pulmonary Disease (COPD) is a major cause of mortality worldwide, ranking 4th with approximately 5\% of all global deaths \cite{whocopd}. Moreover, the global burden of COPD has been increasing since the 1990s, with current projections approaching 600 million patients with COPD globally by 2050 \cite{wang2025global, boers2023global}. 

The natural evolution of COPD is characterized by episodes of acute symptomatic and functional deterioration \cite{wedzicha2006impact}.  These episodes of exacerbation are among the major causes of COPD-related hospitalizations, which account directly or indirectly for about 10\% of all hospitalizations \cite{Gunen2005}. COPD-related hospitalizations have a huge impact on patients' lives, with an in-hospital mortality rate ranging from 2.5\% to 11.5\%, a one-year post-discharge mortality between 9.8\% and 23\%, and readmission rates at one year from discharge from 6.7\% to 35.1\% \cite{waeijen2024global}. 

Survival analysis refers to methods designed to analyze time-to-event data, where the primary outcome is the time until the occurrence of specific events of interest. Unlike traditional regression techniques, survival analysis techniques are able to deal with data that can be partially censored, truncated, or both, making survival analysis particularly valuable in clinical and epidemiological research \cite{wiegrebe2024deep}. Traditional statistical approaches include both non-parametric methods, such as the Kaplan-Meier estimator \cite{kaplan1958nonparametric}, and semi-parametric methods, in particular the Cox proportional hazards regression model and its variations \cite{cox1972}. However, these models often underperform with high-dimensional or time-varying data and may fail to capture complex non-linear dependencies. For this reason, in recent years machine learning (ML) methods first, and then deep learning (DL) methods, have emerged, leveraging a higher representational power to capture intricate patterns in the inputs, improving the predictive performance in time-to-event modeling \cite{wiegrebe2024deep}. Existing prognostic models for COPD outcome prediction focus mainly on mortality prediction, followed by exacerbation and hospitalization predictions. The most common predictors for these models are age, sex, smoking status, Body Mass Index (BMI), Forced Expiratory Volume in 1 second (FEV$_1$) \cite{bellou2019prognostic}. A prospective observational study in five university hospitals in Sweden, Denmark, and Iceland, for instance, fitted a Cox model on 256 patients followed for almost 9 years, finding associations between patient mortality and various predictors, including age, diabetes history, low BMI, and low FEV$_1$ values \cite{gudmundsson2012long}. In a more recent study, the effects of different predictors on mortality of COPD and asthma patients were investigated on a cohort of 5,922 Finnish patients with an 18-year follow-up period. Specifically, a Cox model was used, adjusting for sex, age, smoking, education level, BMI, physical activity, and comorbidities \cite{mattila2023mortality}. In another retrospective cohort study, data from 37,938 patients from the Brazilian National Health System were used to fit a Cox model with a 13-year follow-up. In this case, the predictors used include socio-demographic characteristics, BMI, previous diagnosis of asthma and lung cancer, and therapeutic regimens \cite{Gargano2022}.

Regarding the prediction of hospitalizations and/or exacerbations, the ACCEPT predictive tool \cite{adibi2020accept} used data of 2,380 patients to develop a mixed model to predict both failure time and severity of COPD exacerbations (mild for exacerbations requiring only short-acting bronchodilators, moderate for exacerbations requiring systemic corticosteroids and/or antibiotics, and severe for exacerbations requiring hospitalization), based on history of exacerbations, age, sex, BMI, smoking status, domiciliary oxygen therapy, lung function, symptom burden, and medications. In another study, data of 503 patients from the Sherbrooke University Hospital were used to develop a Cox-based predictive model for COPD failure (defined as both hospitalization and death), using a Dirichlet-based weighting scheme to automatically down-weight weak or noisy predictors \cite{zhang2019feature}.

Machine learning and deep learning approaches could potentially improve predictive performance in these studies. Application to COPD outcomes, however, is still limited, even if growing. A recent review found only 18 studies that applied ML and DL models for the prediction of COPD-related outcomes, with only 6 focused on DL \cite{smith2023machine}. Of these studies, only a few used ML or DL for survival analysis: Moll et al. \cite{Moll2020} used Random Survival Forests (RSF) \cite{ishwaran2008random} to select important features from 30 input variables—a combination of clinical, spirometric, and CT-based imaging features—used to train a Cox model to predict all-cause mortality in COPD patients. Nam et al. \cite{nam2022deep}, on the other hand, used a convolutional neural network (CNN) survival prediction model, using chest radiographs alone and with added clinical variables to forecast 5-year survival in COPD patients. A similar study \cite{yun2021deep} used CT scans and CNN, outperforming classical quantitative CT markers for survival prediction in COPD. Moreover, the results of these reviews suggest that ML/DL models do not consistently perform better than conventional regression models \cite{smith2023machine, Christodoulou2019}.

In this study, we performed survival analysis to predict two events in COPD patients, namely hospitalization and death, using longitudinal features extracted from  Electronic Health Records (EHRs). Specifically, we tested different linear, ML and deep survival analysis techniques, to assess how different losses and architectures performed in the prediction of the two outcomes. To the best of our knowledge, this is the first work focused on time-to-event prediction of multiple events in COPD patients, while two previous works tried to predict exacerbations and mortality but not in a time-to-event framework and using CT images rather than  EHR data \cite{singla2021improving, gonzalez2018disease}.

    
    
%
    

    
\section{Study Design and Methods}
\subsection{Data source and study population}\label{sec:datasource}
Data for this study were extracted from the Information System for the Development of Research in Primary Care (SIDIAP) database \cite{Recalde2022}. SIDIAP is a population-based electronic health record database designed to facilitate epidemiological and clinical research in Catalonia, Spain. It compiles anonymized data routinely collected from primary care practices managed by the Catalan Health Institute (Institut Català de la Salut, ICS), which serves approximately 80\% of the region’s population. The database includes information on around 5.6 million individuals and contains detailed records on patient demographics, clinical diagnoses, visits to healthcare professionals, laboratory test results, prescribed and dispensed medications, and various clinical and lifestyle variables, reflecting real-world clinical practice.

We extracted data from the period between January 1, 2013 and December 31, 2017 for patients with an active diagnosis of COPD, defined by codes from group J44 in ICD-10 \cite{Bramer1988}. Subjects were excluded if they were under 18 years of age at the beginning of the study period. For each patient, the extracted data included clinical and lifestyle variables such as diastolic and systolic blood pressure (DBP and SBP), weight, body mass index (BMI), total cholesterol, blood glucose levels, smoking status (smoker, former smoker, non-smoker), self-reported alcohol consumption and risk of alcoholism, weekly physical activity, Medical Research Council (MRC) dyspnoea scale, spirometry results, and electrocardiogram (ECG) measurements. Additionally, we collected data on prescribed medications for COPD treatment, corresponding to codes R01–R03 in the Anatomical Therapeutic Chemical Classification System, 7th edition (ATC-7) \todo{no entiendo que farmacos tenemos, en el protocolo pone agruopador APREMPOC sin mas}\cite{WHO2021}. Data on hospital discharges (as ICD-9 codes) and diagnoses of chronic lower respiratory diseases (ICD-10 codes J40–J44) were also included.

\subsection{Study design and outcomes definition}
\begin{figure}
    \centering
    \includegraphics[width=0.95\linewidth]{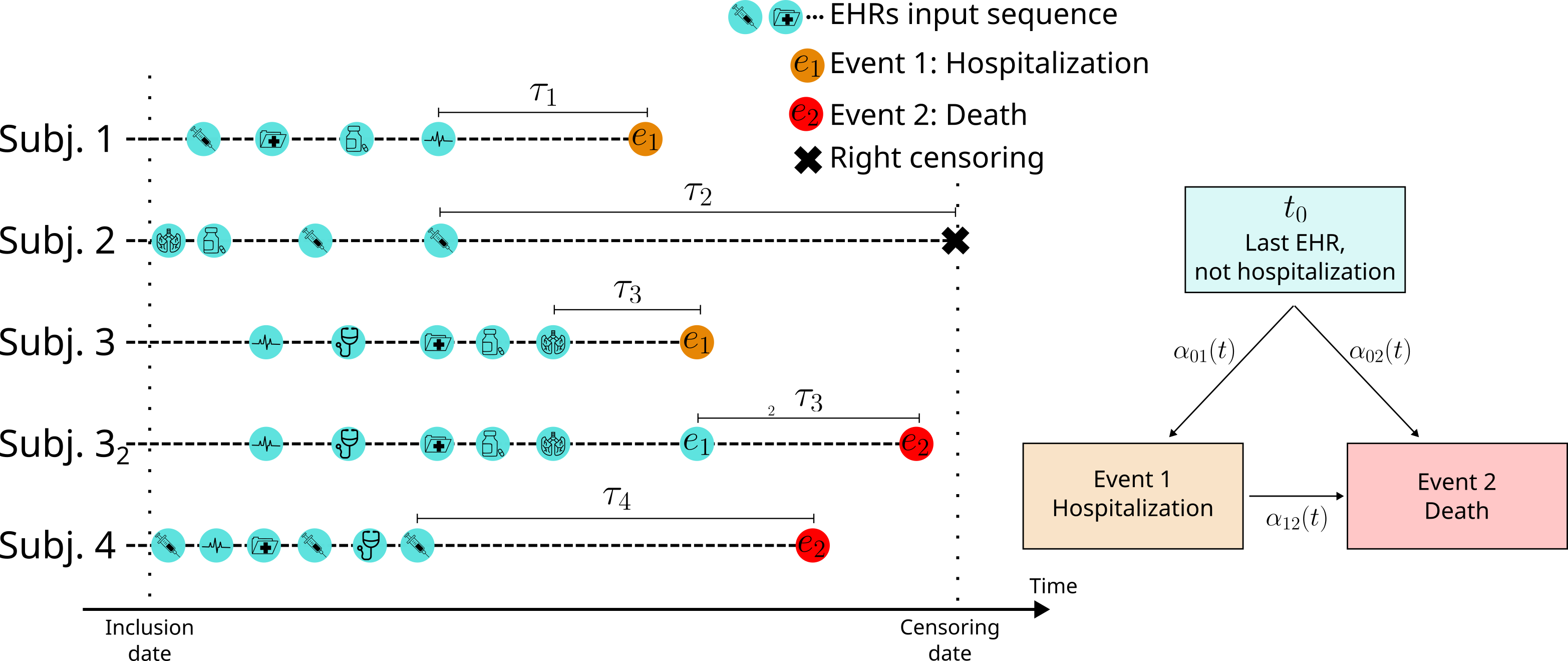}
    \caption{Schematic representation of survival data from longitudinal EHRs (left) and the transition model for semi-competing events (right). The models aim to predict the cause-specific hazard functions $\alpha_{ij}(t)$ from the input longitudinal EHRs. Patients can suffer one of the two events (Subject 1 and 4) or be right censored (subject 2). Moreover, subjects can experience event 2 after event 1 (Subject 3). In this case, the first event is considered in the input sequence when predicting the second event. }
    \label{fig:inps}
\end{figure}
The objective of the present study is to estimate the probability distribution of the first hitting time for two specific semi-competing risks using deep learning models applied to longitudinal sequences  of EHRs. The two events of interest are the first hospitalization occurring after the last record in the input sequence and death, occurring either after the last record in the input sequence or after the first event (see Figure \ref{fig:inps}, right). Patients who experienced neither event by the end of the observation period (2017-12-31) and patients who were transferred during the period were treated as right-censored. Moreover, as most of the models were not specifically designed for a semi-competing scenario, we also evaluated all models under a full competing risks setting, i.e. calculating only the two hazard $\alpha_{01}$ and $\alpha_{02}$ in the transition model in Figure \ref{fig:inps}.

For each patient, the input sequence consisted of EHRs recorded from the inclusion date (2013-01-01) up to the last non-hospitalization EHR prior to 2017-12-31 (see Figure \ref{fig:inps}). To reduce data sparsity, EHRs were aggregated by month; when the same variable was recorded multiple times in a month we used the mean, and we applied mean imputation to handle missing values. Additionally, we excluded patients with fewer than four records in the input sequence, and sequences longer than 24 months were truncated to include only the last two years of records.

The full sequence of EHR data was used as input for models capable of handling longitudinal inputs. Continuous variables were scaled before being input into the model. For models requiring static inputs, only the last available measurement of each variable in the observational period was used.

\subsection{Models}
To predict the two competing outcomes, we tested a range of models spanning statistical methods, machine learning, and deep learning approaches.

\paragraph{Statistical Models} We included the cause-specific Cox proportional hazards model (CS-COX)\cite{cox1972} and the Fine-Gray subdistribution hazards model\cite{fine1999proportional}, both of which are standard techniques for competing risks analysis \cite{Beyersmann2012}. For both models, we used all available input variables. However, ICD codes for diagnosis and ATC codes for drugs where grouped by chapter, to reduce input dimensionality and reduce correlation in case of multiple codes the same chapter.  

\paragraph{ML Models} As an ML approach, we evaluated the SurvivalBoost \cite{alberge2024}, a model that estimates the cause-specific Cumulative Incidence Function (CIF) for each event of interest using a Gradient Boosting Decision Tree classifier.

\paragraph{DL Models with static inputs} We tested two models inputs: SurvTRACE \cite{wang2022survtrace} leverages transformers for survival analysis with competing risks handling categorical and continuous input variables with two specific encoders; and DeepPseudo \cite{rahman2021deeppseudo}, a model that leverages pseudo-values to reduce a complex survival analysis to a standard regression problem.

\paragraph{DL Models with longitudinal inputs}To the best of our knowledge, there are only two models designed for predicting competing events from multivariate time series: Dynamic-DeepHit (DDH)\cite{lee2019dynamic}, a model designed to handle longitudinal inputs and dynamic prediction of competing events leveraging recurrent neural networks (RNNs) and an attention mechanism; and Deep Recurrent Survival Machine (DRSM)\cite{nagpal2021deep}, which combines RNNs with mixture density estimation to model survival distributions over time. 

Specific details on the models functioning and implementation can be found in Appendix 1.

\subsection{Statistical analysis and model evaluation}
Patients' characteristics were reported using mean and SD for continuous variables, and percentages for categorical variables. Model comparisons were performed with 10-fold cross-validation. Deep Learning models have been trained for a maximum of 70 epochs with early stopping.  We employed the Wilcoxon signed-rank test ($\alpha=0.05$) to compare the performances of the different models, evaluating using the concordance index and the cumulative dynamic area under curve (Cumulative-AUC) at every month of the follow-up period.

For DL models,we searched for the best hyperparameters configuration using Bayesian Optimization HyperBand algorithm \cite{falkner2018bohb}. All models were been implemented with PyTorch \cite{paszke2017} library. Statistical models (CS-COX and Fine-Gray) were implemented in R (4.1.2). 

\section{Results} 

\subsection{Study participants}
\begin{figure}[ht]
  \centering
  \begin{subfigure}{.9\textwidth}
    \centering
    \includegraphics[width=\linewidth]{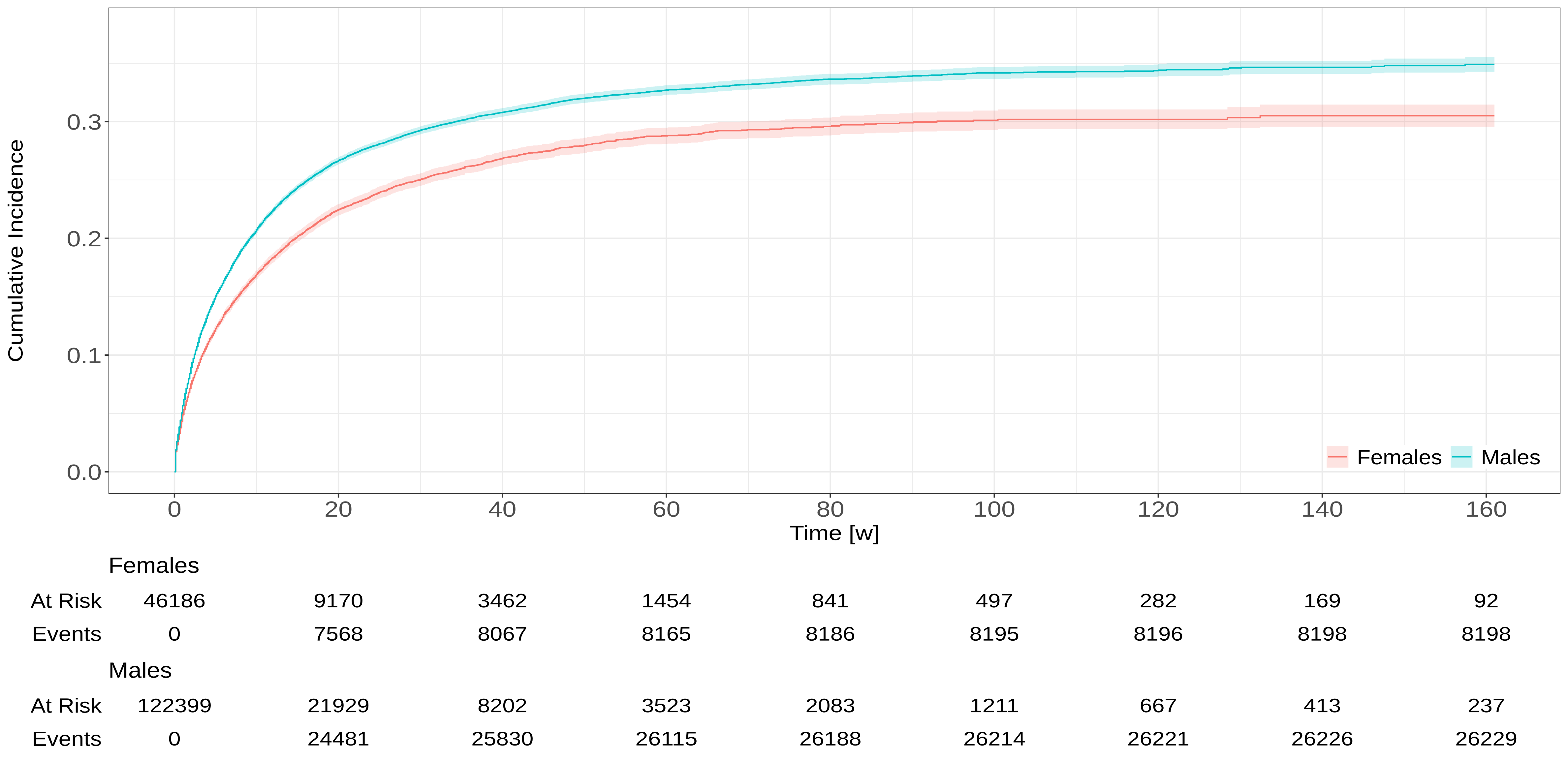}
    \caption{Hospitalization}
    \label{fig:sub1}
  \end{subfigure}%
  \\
  \vspace{0.3cm}
  \begin{subfigure}{.9\textwidth}
    \centering
    \includegraphics[width=\linewidth]{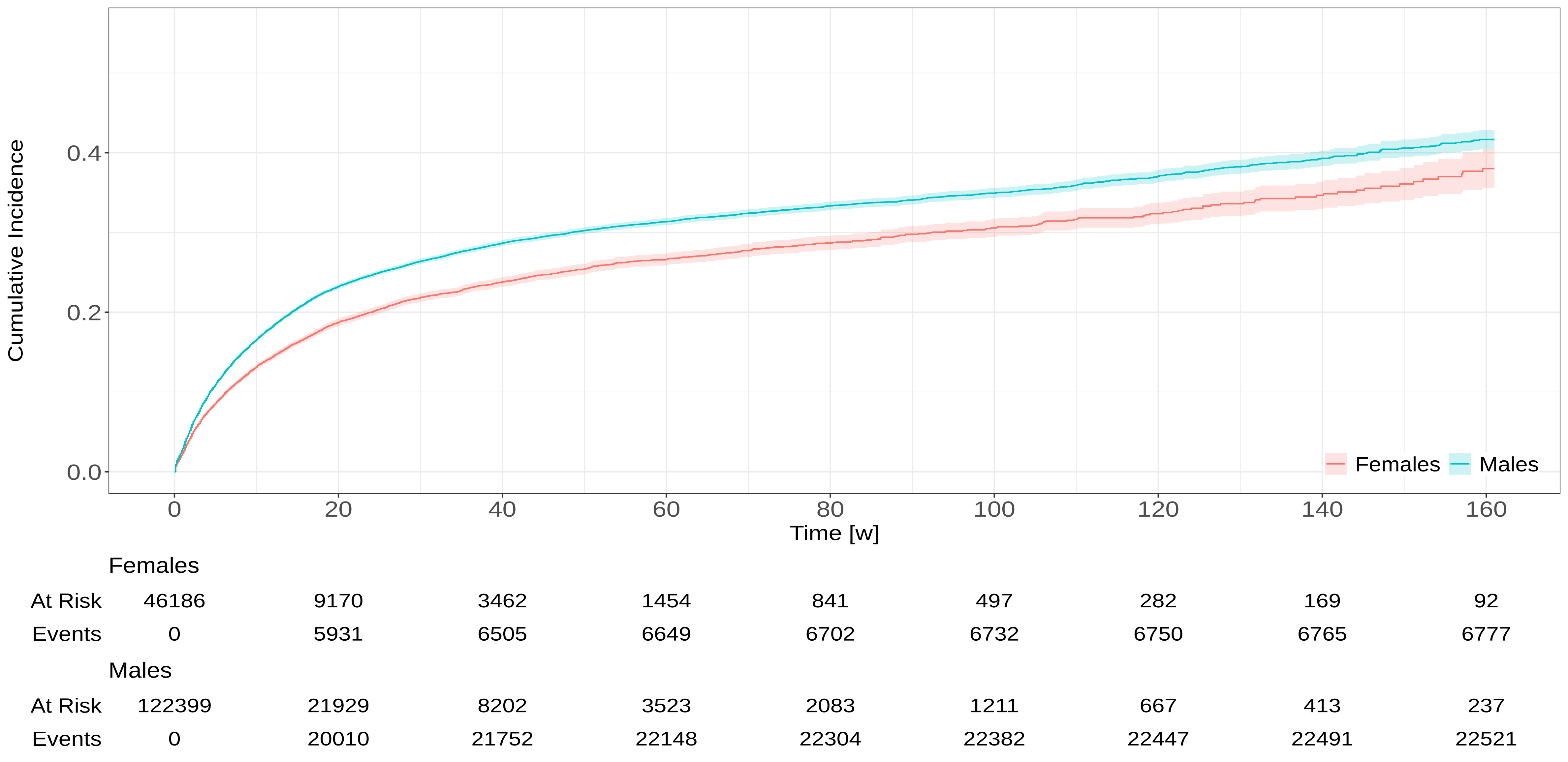}
    \caption{Death, all}
    \label{fig:sub2}
  \end{subfigure}
  \begin{subfigure}{.9\textwidth}
    \centering
    \includegraphics[width=\linewidth]{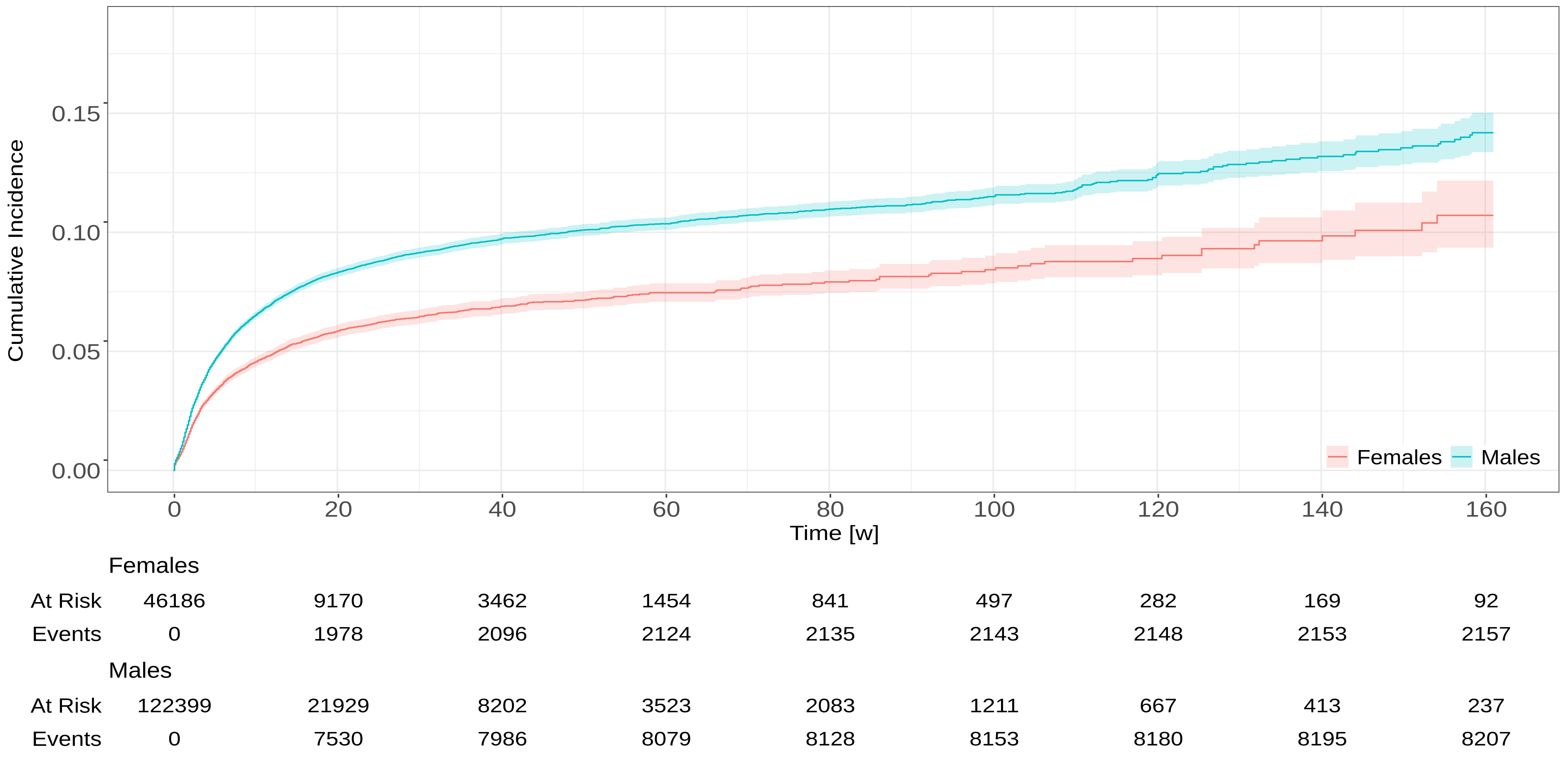}
    \caption{Death after hospitalization}
    \label{fig:sub3}
  \end{subfigure}
  \caption{Aalen–Johansen estimator of the cumulative incidence functions for the two events, stratified by sex.}
  \label{fig:ci}
\end{figure}

Data of 220,657 subject with COPD were extracted from the SIDIAP database. After applying the selection criteria described in section \ref{sec:datasource}, we obtained 158,204 subjects used to train the models, of whom 114,177 (72.17\%) males. Mean age at inclusion (2013-01-01) was $69.21$ years ($SD=12.45$). Among the 62,453 excluded subjects, 76.12\%were males and mean age at inclusion was $78.66(SD=14.29)$. Mean number of records for subject was $82(SD=53)$. The baseline characteristics of the records are reported in Appendix 3, Table \ref{tab:baselines}. The mean follow-up time was $12.72$ weeks ($SD=19.44$).  Aalen-Johansen estimate \cite{aalen1978empirical} for the cumulative incidence functions of the two events of interest are shown in Figure \ref{fig:ci}: in total 34431(21.76\%) subjects were hospitalized during the follow-up, with a mean time-to-event of 6.43(9.83) weeks; 29345(18.55\%) subjects died during follow-up, with a mean time-to-event of 9.67(16.60) weeks; 10381(6.43\% of the total, 30.15\% of hospitalizations) subjects died after the last hospitalization,  with a mean time-to-event of 8.11(16.00) weeks, while 18964 subjects did not have hospitalization records between inclusion date and date of death. 

\subsection{Models performances}

Results of the hyperparameters search are reported in Appendix 3, Tables \ref{tab:surv}-\ref{tab:drsm}. Table \ref{tab:results}(a) presents the performance of the models for the two events of interest using 10-fold cross-validation in the semi-competing events scenario. Overall, DDH and RDSM demonstrate superior predictive capabilities across both tasks. Specifically, in terms of Concordance Index, DDH achieves the highest score for hospitalization (0.81), indicating excellent discriminative ability in predicting event order. RDSM also performs well (0.79), while other models appear to struggle with hospitalization prediction. For the death event, RDSM attains the highest concordance (0.84), closely followed by DDH (0.83).

Regarding Cumulative-AUC, DDH  outperforms all other models for hospitalization (0.83), with SurvTRACE (0.81) and RDSM (0.80) also showing competitive performance. For the death task, SurvTRACE achieves the highest AUC (0.85), followed by RDSM (0.82) and DDH (0.80). 

Similar results were obtained also for the competing risks model (Table \ref{tab:results}(b)): DDH achieved the best metrics across both events, followed by RDSM. Also, SurvTrace model maintains good performance in terms of Dynamic-AUC.  

In general, traditional models such as CS-COX and FG show moderate performance, yielding better results for death prediction but struggling with hospitalization. In contrast, DeepPseudo and  SurvivalBoost generally underperform across both metrics and tasks, with better results in the competing risks scenario, while both models struggle to predict the two events in the semi-competing risks configuration.

To asses variables relevance in the outcomes prediction, we estimated SHAP values using the integrated gradients method \cite{lundberg2017unified} for the top performing model (Figure \ref{fig:shaps}).  Hazard ratios for the CS-COX model are reported in Appendix 2, Figure \ref{fig:hr_cox}. 

\begin{table}[ht]
\caption{\textbf{Models results in 10-fold cross validation}, mean value(std). Bold indicates the highest value; underline indicates the second highest. $p$ values refer to tests between the highest and second-highest models.  ($^{***}$ for $p<0.001$, $^*$ for $0.01<p<0.05$, $^.$ for $0.05\leq p <0.1$, Wilcoxon signed-rank test).}
\centering
\caption*{(a) Semi-competing Risks}
\footnotesize
\begin{tabular}{lcc|cc}
    \toprule
    & \multicolumn{2}{c|}{Concordance Index} & \multicolumn{2}{c}{Cumulative-AUC}\\
      & Hospitalization & Death & Hospitalization & Death \\ 

    \midrule
    CS-COX     & 0.65(0.02)   & 0.81(0.03)     & 0.71(0.01)   & 0.80(0.05)  \\   
    Fine-Gray  & 0.65(0.02)   & 0.81(0.03)     & 0.70(0.01)   & 0.78(0.05)  \\         
    DDH           & \textbf{0.81(0.01)}$^{***}$   & \underline{0.83(0.02)}     & \textbf{0.83(0.01)}   & 0.80(0.05)  \\
    SurvTRACE            & 0.66(0.02)   & 0.79(0.01)     & \underline{0.81(0.02)}   & \textbf{0.85(0.02)}$^{***}$   \\
    RDSM          & \underline{0.79(0.02)}   & \textbf{0.84(0.02)}$^.$     & 0.80(0.01)   & \underline{0.82(0.04)}  \\
    DeepPseudo    & 0.61(0.04)   & 0.55(0.02)     & 0.63(0.01)   & 0.67(0.05)  \\
    SurvivalBoost & 0.56(0.01)   & 0.81(0.02)     & 0.70(0.01)   & 0.78(0.04)  \\
    \bottomrule
\end{tabular}
\vspace{1em}
\caption*{(b) Competing Risks}
\footnotesize
\begin{tabular}{lcc|cc}
    \toprule
    & \multicolumn{2}{c|}{Concordance Index} & \multicolumn{2}{c}{Cumulative-AUC}\\
      & Hospitalization & Death & Hospitalization & Death \\ 

    \midrule
    CS-COX     & 0.68(0.02)   & 0.80(0.02)     & 0.73(0.00)   & 0.81(0.02)  \\   
    Fine-Gray        & 0.68(0.02)   & 0.80(0.02)     & 0.73(0.00)   & 0.80(0.02)  \\    
    DDH           & \textbf{0.86(0.01)}$^{*}$   & \textbf{0.85(0.03)}     & \textbf{0.88(0.00)}$^{***}$   & \textbf{0.84(0.02)}  \\
    SurvTRACE            & 0.69(0.01)   & 0.79(0.02)     & 0.83(0.02)   & \textbf{0.84(0.05)}   \\
    RDSM          & \underline{0.84(0.02)}   & \underline{0.84(0.02)}     & \underline{0.84(0.00)}   & \underline{0.83(0.03)}  \\
    DeepPseudo    & 0.61(0.04)   & 0.69(0.04)     & 0.70(0.02)   & 0.75(0.03)  \\
    SurvivalBoost & 0.62(0.01)   & 0.81(0.02)     & 0.73(0.00)   & 0.81(0.02)  \\
    \bottomrule
\end{tabular}
\label{tab:results}
\end{table}


\begin{figure}[ht]
    \centering
    \includegraphics[width=0.9\textwidth]{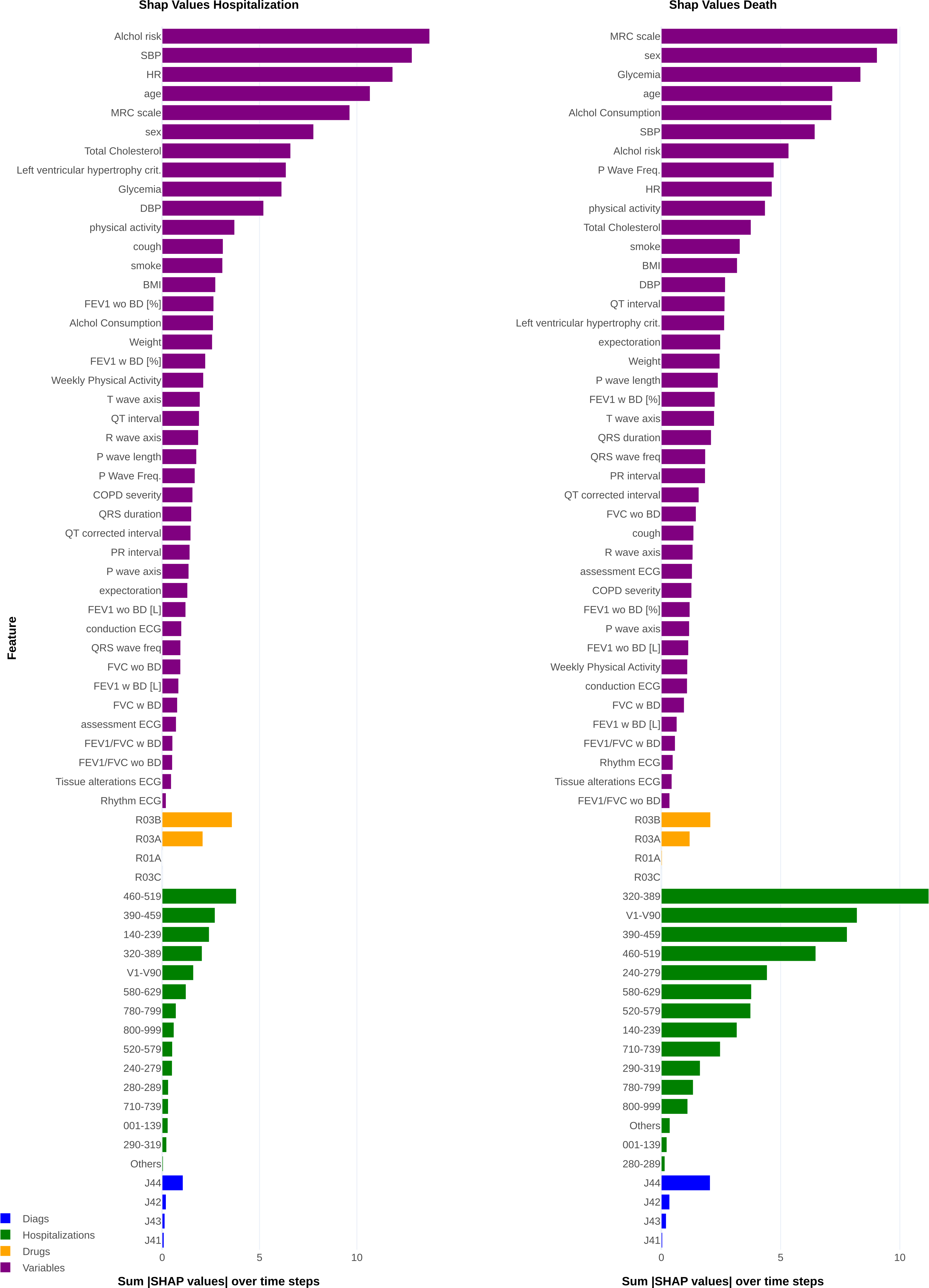}
    \caption{Sum of the absolute values of the SHAP estimation coefficients over time steps for the DDH model, weighted by the time attention of the model.}
    \label{fig:shaps}
\end{figure}

\section{Discussion}
 COPD hospitalizations due to acute exacerbations are a burden for health systems, with a cost that can be up to 60 times higher than mild or moderate exacerbations, i.e., those exacerbations self-managed by the subjects or that just require telephone contact with a health-care center \cite{andersson2002costs}. Hence, understanding the likelihood of future hospitalizations in COPD patients is crucial in the management of the disease \cite{kaleem2022impact}. By identifying high-risk patients, personalized ML and DL based models could enable targeted interventions that reduce hospitalizations and healthcare costs. The impact of these model on COPD-related outcomes is, however, unclear \cite{smith2023machine}.

Hence, the objective of this study is to compare statistical, ML and DL models for time-to-events predictions in COPD patients. Specifically, sequences of EHRs were used to predict time to the first hospitalization after the last record, with death treated as a semi-competing event. While mortality risk and the prediction of hospitalization or acute exacerbations are the most commonly studied outcomes for ML and DL models \cite{smith2023machine}, only a few studies have simultaneously focused on predicting both outcomes, typically using CT images as model inputs. To the best of our knowledge, this is the first study to develop predictive models for hospitalization and death using longitudinal EHR data.

To train the models we extracted data of 220657 subjects from the SIDIAP database, corresponding to the whole COPD population in Catalonia (Spain). After applying the selection criteria, we obtained data of 158204 subjects. Approximately 22\% of subjects suffered an hospitalization during the follow-up. This percentage is consistent with other studies, where it has been found that between 8 and 20\% of COPD patients require at least one hospitalization per year, while 35\% of them have been hospitalized at least once \cite{westbroek2020reducing, izquierdo2021clinical}. 19\% of subjects died during the follow-up. This percentage is higher between hospitalized patients (30\%), consistent with other studies that found that mortality rate after hospital admission are up to 43\% \cite{westbroek2020reducing}. 

A recent review suggests that ML and DL models show minimal or no improvements compared to linear methods for predicting COPD-related outcomes \cite{smith2023machine}. Another review, centred on more general clinical prediction models, found no evidence of benefits from ML models over linear regression models \cite{Christodoulou2019}. Our results partially confirm these findings in a survival task scenario. Specifically, the SurvivalBoost model and the DeepPseudo model demonstrated lower performance compared to the Cox-based models, with the DeepPseudo model showing the worst metrics overall, and the SurvivalBoost model matching the prediction of the second event but struggling with the first. The SurvTrace model, on the other hand, demonstrates competitive or superior performance, especially in terms of Cumulative-AUC. SurvTRACE achieves the highest AUC for death prediction in both scenarios, suggesting its strength in capturing temporal risk profiles.

The most significant improvements come from models that incorporate recurrent inputs. These models outperform all others in terms of concordance and show strong AUC values across both events, particularly for hospitalization, which appears to be the more difficult outcome to predict. This suggests that temporal modeling of longitudinal patient information—enabled by recurrent structures-plays a critical role in improving predictive performance, especially for dynamic and less deterministic events like hospitalization. In contrast, simpler models that do not leverage sequential patterns tend to underperform across both tasks, further supporting the importance of capturing temporal dependencies in survival modelling.

Several variables showed high relevance in predicting both outcomes, including sex, age, alcohol consumption, heart rate, and the MRC scale. Among these, alcohol use and the MRC scale also emerge as important predictors in the CS-COX model, highlighting their consistent role across modeling approaches and confirming their predictive power \cite{tsimogianni2009predictors, alqahtani2020risk}. Moreover, being female has already been identified as a protective factor for both hospital admission \cite{alqahtani2020risk} and mortality \cite{owusuaa2022predictors}. Also elevated glycemia levels have  been shown to independently predict severe exacerbations and short-term mortality in COPD, reinforcing the importance of metabolic status in risk stratification \cite{chen2022elevated, ali2025glycated}. In addition, a history of previous hospitalizations shows high SHAP values, particularly for diagnostic groups 320–389 (diseases of the nervous system and sense organs), 390–459 (diseases of the circulatory system), and 460–519 (diseases of the respiratory system). For both outcomes, high SHAP values are also present for additional factors influencing health status (V01–V90), which accompany the main hospitalization code.

Nonetheless, it is important to emphasize that the aim of this work is to compare models in their ability to predict survival outcomes, rather than to establish causality or study the direct relationships between individual variables and outcomes. The variable importance rankings should therefore be interpreted with caution, as they serve primarily as indicators of which features contribute most to prediction within each framework.

Moreover, the models were evaluated using internal 10-fold cross-validation, but were not externally validated on independent datasets. This limits the generalizability of our findings across different healthcare settings and populations. Finally, although we explored a variety of modeling strategies, the inclusion of pretrained transformer-based architectures for EHR data—an emerging direction in DL research—remains an area for future work.

\todo[inline]{Other comments that can be added here but we reached CHEST max number of words (3200): 1. loss of survivalboost is weighted by and estimation of the censoring probability, that here is quite higher then the ds tested in the paper, can be an expalnation of the bad results. 2. Survtrace is based on transformer, it could be it has not enough data to be trained 3. Deeppseudo fails also because pseudo values estimation is somehow failing 4. even with similar metrics RDSM is quite slower to be trained}

\section{Conclusions}

In this study, we conducted both statistical and deep time-to-event analyses for two clinically relevant outcomes in COPD patients—hospitalization and death—using real-world data from the SIDIAP database. While previous reviews have questioned the added value of ML and DL approaches in clinical prediction, our findings offer a more nuanced perspective. Linear models such as CS-COX and Fine-Gray demonstrated moderate performance, particularly struggling with the prediction of hospitalizations. In contrast, more advanced models showed improved time-dependent discrimination. The best overall performance was achieved by models that incorporate recurrent inputs, specifically DDH and RDSM. These models consistently outperformed others in terms of both Concordance Index and Cumulative-AUC, with particularly significant gains in predicting hospitalization. These results underscore the importance of capturing temporal patterns in longitudinal EHRs for effective survival prediction.

\subsection*{Acknowledgment}
This work was supported by the Grant PID2021-122952OB-I00 funded by AEI 10.13039/501100011033 and by ERDF A way of making Europe; the Networking Biomedical Research Centre in the subject area of Bioengineering, Biomaterials and Nanomedicine (CIBER-BBN), initiatives of Instituto de Investigación Carlos III (ISCIII); ISCIII (grant AC22/00035); and the CERCA Programme / Generalitat de Catalunya. B2SLab is certified as 2021 SGR 01052. 

\FloatBarrier
\bibliography{refs}

\section*{Appendix 1: Models Functioning and Implementation Details}

This section provides additional details on the machine learning and deep learning models evaluated in our study, with a focus on their architecture, model-specific input processing, survival prediction mechanisms, and implementation details. For each model, we also report the hyperparameter space. When clearly specified, we explored the same hyperparameter space used by the original authors. However, to reduce the search space and consequently the training time, we fixed the learning rate to $1 \times 10^{-4}$ and the batch size to 256, unless otherwise stated.

\paragraph{SurvivalBoost}

SurvivalBoost \cite{alberge2024} is a gradient boosting-based model specifically developed for survival analysis in the presence of competing risks and right censoring, grounded in a strictly proper scoring rule framework. The model estimates the cause-specific cumulative incidence functions (CIFs) by optimizing a novel loss function that is both separable (i.e., decomposable over individual observations and time points) and censoring-aware, enabling efficient stochastic optimization (e.g., via mini-batch training). The prediction target is the discretized CIF for each event type $c$ over a predefined time grid $\{t_1,...,t_K\}$. For each event cause, SurvivalBoost fits a separate Gradient Boosted Decision Tree (GBDT) model to estimate $\hat{F}(t_k|x)$, i.e., the probability of the event occurring by time $t_k$, conditioned on covariates $x$.

The loss function used for optimization is based on a reweighted version of a log loss, with weights derived from the inverse probability of censoring. Specifically, each observation's contribution at time $t_k$ is weighted by $1/\hat{G}(t_k|x)$, where $\hat{G}$ is an estimate of the probability of remaining censor-free at time $t_k$. The authors demonstrated that this reweighting constitutes a strictly proper scoring rule for the global CIF on a fixed time horizon.


In order to test the SurvivalBoost model, we used the \href{https://github.com/juAlberge/hazardous}{hazardous repository} provided by the authors. Main hyperparameters explored were: learning rate (0.01, 0.05, 0.1); max leaf nodes (10, 20, 30, 40, 50); min samples leaf (10, 30, 50, 70, 90, 110); and number of iterations (75, 100, 150, 200). Note that, differently from the deep learning models that are trained through different epochs, SurvivalBoost is a tree-based method. Hence, for the search of the optimal hyperparameters, we used a simple Random Search, as proposed by the authors.

\paragraph{Deeppseudo}
DeepPseudo\cite{rahman2021deeppseudo} reformulates competing-risk survival analysis as a regression problem using pseudo‐values derived from nonparametric Aalen–Johansen estimators to handle censoring. The model predicts the cause-specific cumulative incidence functions (CIFs) at predefined time horizons by estimating subject-specific pseudo‐values for each cause and time.

Specifically, as the original work proposed four variations of the model, we used the Cause-specific Marginal DeepPseudo Model that estimates the CIF using Jackknife pseudo-values, based on leave‐one‐out Aalen–Johansen estimates of CIF, yielding an unbiased "target" even under right‐censoring. We chose this implementation mainly for the faster calculation of the pseudo-values with the Jackknife function. The model is then composed of a shared sub-network based on a multilayer perceptron to learn the shared representation of the competing events and $n$ cause-specific sub-networks used to predict the pseudo-values (and thus the marginal CIFs) for a specific cause at $M$ evaluation time points. This architecture is inspired by the DeepHit model\cite{lee2018deephit}. The model is trained with a Mean Squared Error loss between predicted and target pseudo‐values across all causes and time points.

We used the functions provided by the authors in the \href{http://github.com/umbc-sanjaylab/DeepPseudo_AAAI2021/blob/main/CS_Marginal_DeepPseudo/Compute_pseudo_values.r}{deeppseudo repository} to estimate the pseudo-values from our data, while the model, composed of standard MLP and loss functions, was implemented in PyTorch, for coherence with the other models. The hyperparameter space was defined by the same parameters proposed by the authors (number of layers  and number of neurons in the cause-specific and shared networks (1 to 5-50, 100, 200, 300 respectively), batch size (64, 128, 256), plus dropout (0, 0.1, 0.3, 0.6), which was not included in the original work.

\paragraph{SurvTRACE} SurvTRACE \cite{wang2022survtrace}  is a deep learning architecture based on transformers, explicitly designed for competing risks survival analysis with right-censored data. Differently from the original transformer, the embedding of the model is composed of two parallel embedding layers: discrete features are transformed into one-hot vectors and mapped into the embedding space with an embedding matrix, while numerical features are passed through a dense layer to be mapped into the same embedding space. These two types of embeddings are concatenated before being passed to the attentive encoder layers, where covariate embeddings interact to form high-order combinatorial embeddings. The obtained representations are fed into event-specific output heads. Each head outputs a discrete-time hazard function over predefined time intervals for one event type. Hazard outputs are aggregated to form the predicted CIFs. In this way, the model avoids any parametric assumptions about underlying survival distributions, shared across all task-specific sub-networks for downstream tasks. The primary loss is an IPS-weighted negative log-likelihood over observed events and time intervals \cite{kvamme2021continuous}. The authors suggested that auxiliary losses could improve the model’s performance; however, these were not implemented in the \href{https://github.com/RyanWangZf/SurvTRACE/blob/main/survtrace/losses.py}{SurvTRACE repository}, hence the model was trained only with the primary survival loss.

Following the authors’ experiments, we searched for the best hyperparameter configuration in a space composed of the number of transformer layers (1,2,3), the embedding size (12, 24, 48, 96), the intermediate layer size (16, 32, 64), and the number of attention heads (3,6,12).

\paragraph{Dynamic-DeepHit}
Dynamic-DeepHit~\cite{lee2019dynamic} is an extension of the DeepHit~\cite{lee2018deephit} framework tailored to handle longitudinal data for survival analysis with competing risks. The model incorporates dynamic information through recurrent neural networks (RNNs), allowing it to update survival predictions over time as new measurements are observed. At each landmark time $t$, the model takes as input the full longitudinal history up to $t$ and outputs the joint probability of failure from each cause at future time intervals. Specifically, the model is composed of a shared recurrent sub-network (GRU or LSTM) that processes sequential covariates, followed by an attention layer. Then, multiple cause-specific feedforward sub-networks estimate the discrete-time joint distribution $P(T = t_k, \text{event}=c \mid \mathcal{H}_t)$, where $\mathcal{H}_t$ denotes the historical covariate trajectory up to time $t$. The loss function combines a log-likelihood term to encourage correct joint probability estimation and a ranking loss to enforce temporal ordering among failure times. This dual-objective formulation enables the model to capture both the risk at each horizon and the relative ordering of risks across individuals. Moreover, a reconstruction loss is used to predict the next event from the previous one in each RNN layer. 

As the original implementation, available in the \href{https://github.com/chl8856/Dynamic-DeepHit}{Dynamic-DeepHit repository}, used an older version of Python and a different DL framework (TensorFlow 1.3), we implemented the model in PyTorch to test it in the same environment as the other models. Following the authors' experiments, we explored a hyperparameter space that includes the number (1,2,3,4) and type (GRU or LSTM) of RNN layers, the number (1,2,3,4) of hidden units in the attention layer and cause-specific sub-networks and the sizes (100, 200, 300, 350), the dropout rate (0.1, 0.3, 0.6), and the losses parameters $\alpha, \beta, \sigma$ (0.1, 1, 3, 5).

\paragraph{Deep Recurrent Survival Machine}
Deep Recurrent Survival Machine (DRSM)\cite{nagpal2021deep} is a deep parametric survival model designed to handle time-to-event regression with time-varying covariates, combining flexibility of deep learning with interpretability of mixture-based parametric distributions. The model assumes that, conditioned on the covariate history $\mathcal{H}_t$, the time-to-event follows a mixture of $M$ parametric distributions (Weibull or Log-Normal), where the parameters of each component and their mixture weights are dynamically predicted from the observed covariates through a recurrent neural network. Specifically, a shared recurrent encoder (e.g., GRU or LSTM) ingests sequential covariates up to landmark time $t$ and outputs a context vector. This vector is then passed to multiple dense layers to compute the parameters $(\theta_m)$ for each of the $M$ mixture components, as well as their associated mixture probabilities $\pi_m$. The model estimates the conditional survival function $S(t'|\mathcal{H}_t)$ over future time points $t' > t$ using the learned mixture distribution.

To train the model, the authors propose a likelihood-based loss that incorporates both observed and censored instances, enabling the model to learn from right-censored longitudinal data. We used the official implementation available in the \href{https://autonlab.org/auton-survival/models/dsm/index.html}{AutonSurvival package}. However we modified the package to include a dropout, not presented in the original work, and we include the possibility of train the models in the GPUs, as the original package does not include this option. Hyperparameter search included the number of mixture components k (3,4,5,6), type of parametric distribution (Weibull or LogNormal), type (GRU or LSTM), number (1,2,3,4) and dropout (0,0.1,0.3) of RNN layers, hidden units (200, 300, 350, 400), and final layer dropout (0, 0.1, 0.3) .

\newpage
\section*{Appendix 2: Supplementary Images}
\begin{figure}[ht]
    \centering
    \includegraphics[width=\textwidth]{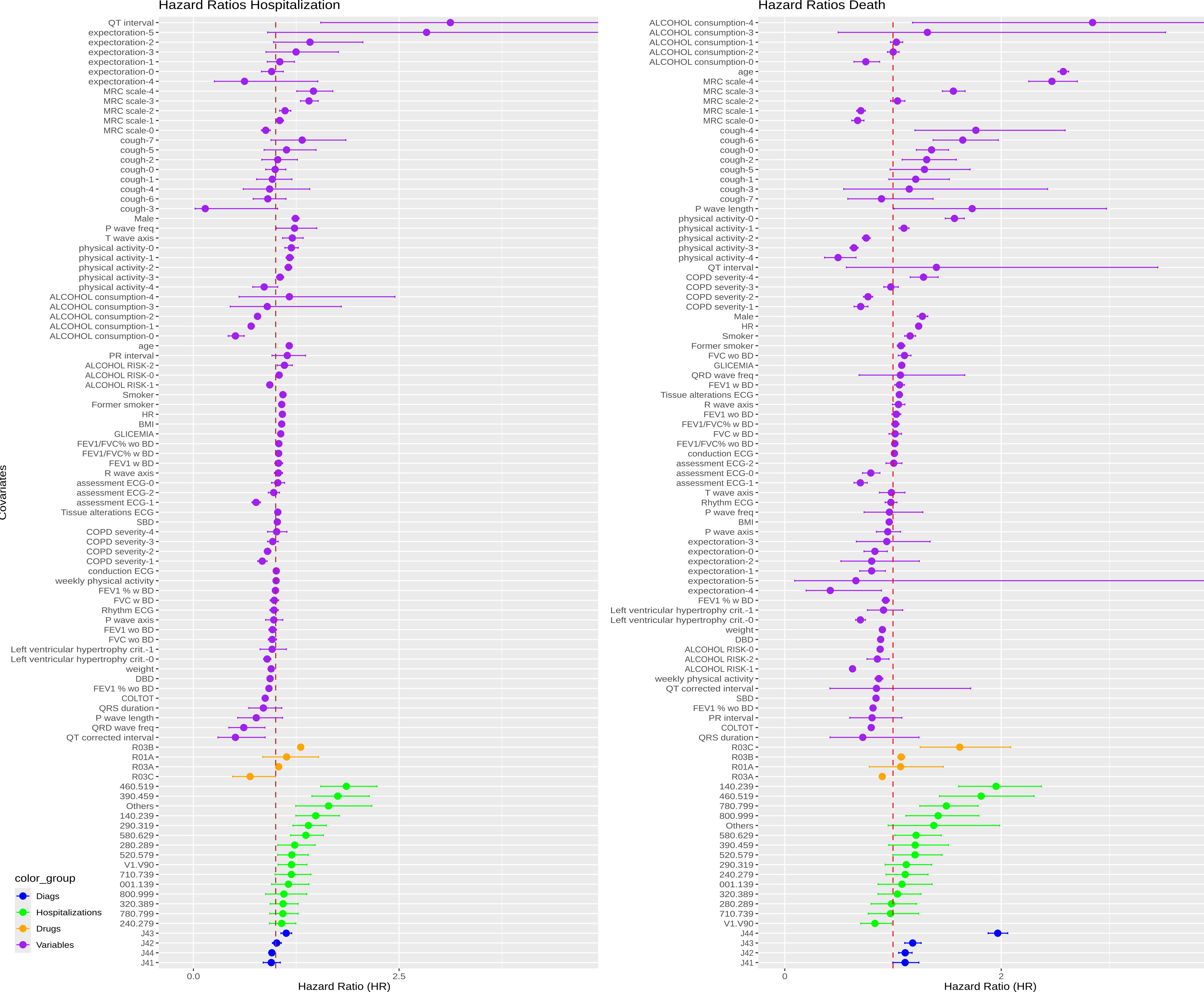}
    \caption{Hazard ratios of the cause specific COX-PH regressor.}
    \label{fig:hr_cox}
\end{figure}

\FloatBarrier

\section*{Appendix 3: Supplementary Tables}
\begin{table}[!ht]
    \caption{\textbf{Baseline characteristics}. Mean value (std) for continuous variables, total count (percentage) for categorical variables and codes.}
    \makebox[\textwidth]{
    \begin{minipage}{.60\textwidth}
        \footnotesize
        \caption*{(a)Variables}
        \centering
        \tiny
        \begin{adjustbox}{width=\textwidth}
        \begin{tabular}[t]{lcc}
        \toprule
         & Value & \# records/pat(std)\\
        \midrule
        \multicolumn{3}{l}{\textbf{Clinical and analytical variables}} \\[1ex] 
       
        Glycemia[mg/dL]  & 112.68(39.77) &  4.62(3.90) \\
        SBP[mmHg]  & 132.64(16.83) &  9.30(8.69) \\
        DBP[mmHg]  & 73.14(10.90) &  9.30(8.69) \\
        HR[bpm]  & 76.04(13.87) &  7.60(8.38) \\
        Weight[Kg]  & 78.73(16.22) &  5.55(5.87) \\
        BMI[kg/m2]  & 29.43(5.38) &  4.66(4.71) \\
        Total Cholesterol[mg/dL]  & 187.69(42.36) &  3.95(2.96) \\
        \midrule
        \multicolumn{3}{l}{\textbf{Lifestyle variables}} \\[1ex] 
        Weekly Physical Activity[h]  & 4.96(3.44) &  0.40(0.96) \\

        Smoke & & 0.57(0.95) \\
        \multicolumn{1}{r}{No} & 44044(27.384) & \\
        \multicolumn{1}{r}{Smoker} & 63372(40.06) & \\
        \multicolumn{1}{r}{Former} & 50788(32.10) & \\
        physical activity & & 1.93(2.42) \\
        \multicolumn{1}{r}{0} & 25949(8.16) & \\
        \multicolumn{1}{r}{1} & 89348(28.09) & \\
        \multicolumn{1}{r}{2} & 106225(33.39) & \\
        \multicolumn{1}{r}{3} & 92433(29.06) & \\
        \multicolumn{1}{r}{4} & 4174(1.31) & \\
        Alcohol risk & & 3.50(2.80) \\
        \multicolumn{1}{r}{0} & 323151(55.19) & \\
        \multicolumn{1}{r}{1} & 240564(41.08) & \\
        \multicolumn{1}{r}{2} & 21818(3.73) & \\
        Alcohol Consumption & & 0.88(1.66) \\
        \multicolumn{1}{r}{0} & 3842(2.70) & \\
        \multicolumn{1}{r}{1} & 57458(40.41) & \\
        \multicolumn{1}{r}{2} & 80386(56.54) & \\
        \multicolumn{1}{r}{3} & 290(0.20) & \\
        \multicolumn{1}{r}{4} & 197(0.14) & \\
        
        \midrule
        \multicolumn{3}{l}{\textbf{Spirometry Results}} \\[1ex] 
        
        FEV1 w BD[L]  & 1.90(0.70) &  0.84(1.04) \\
        FVC w BD[L]  & 2.95(0.94) &  0.83(1.03) \\
        FEV1 w BD[\%]  & 65.10(20.84) &  0.87(1.05) \\
        FEV1 wo BD[\%]  & 61.21(20.04) &  1.16(1.23) \\
        FEV1 wo BD[L]  & 1.77(0.68) &  0.96(1.15) \\
        FVC wo BD[L]  & 2.77(0.92) &  0.95(1.14) \\
        FEV1/FVC w BD[\%]  & 64.38(13.46) &  0.91(1.07) \\
        FEV1/FVC wo BD[\%]  & 64.15(13.96) &  1.17(1.23) \\
        Expectoration & & 0.30(1.57) \\
        \multicolumn{1}{r}{0} & 28873(59.72) & \\
        \multicolumn{1}{r}{1} & 16255(33.62) & \\
        \multicolumn{1}{r}{2} & 1457(3.01) & \\
        \multicolumn{1}{r}{3} & 1417(2.93) & \\
        \multicolumn{1}{r}{4} & 234(0.48) & \\
        \multicolumn{1}{r}{5} & 113(0.23) & \\
        Cough & &0.38(1.69) \\
        \multicolumn{1}{r}{0} & 36169(60.20) & \\
        \multicolumn{1}{r}{1} & 4979(8.29) & \\
        \multicolumn{1}{r}{2} & 7668(12.76) & \\
        \multicolumn{1}{r}{3} & 379(0.63) & \\
        \multicolumn{1}{r}{4} & 1108(1.84) & \\
        \multicolumn{1}{r}{5} & 2419(4.03) & \\
        \multicolumn{1}{r}{6} & 6348(10.57) & \\
        \multicolumn{1}{r}{7} & 1015(1.69) & \\
        MRC scale & & 1.71(2.56) \\
        \multicolumn{1}{r}{0} & 61514(22.11) & \\
        \multicolumn{1}{r}{1} & 123249(44.29) & \\
        \multicolumn{1}{r}{2} & 51321(18.44) & \\
        \multicolumn{1}{r}{3} & 31810(11.43) & \\
        \multicolumn{1}{r}{4} & 10358(3.72) & \\
        COPD severity & & 0.90(1.53) \\
        \multicolumn{1}{r}{1} & 36941(25.47) & \\
        \multicolumn{1}{r}{2} & 70558(48.65) & \\
        \multicolumn{1}{r}{3} & 28642(19.75) & \\
        \multicolumn{1}{r}{4} & 8891(6.13) & \\
        \midrule
        
        \multicolumn{3}{l}{\textbf{ECG assesment}}  \\[1ex] 
        P Wave Freq.[bpm]  & 16.40(51.55) &  0.16(0.47) \\
        QRS wave freq[bpm]  & 77.60(49.76) &  0.17(0.48) \\
        P wave axis[°]  & 50.15(36.16) &  0.15(0.45) \\
        R wave axis[°]  & 12.87(46.68) &  0.27(0.67) \\
        T wave axis[°]  & 45.20(41.49) &  0.17(0.49) \\
        P wave length[ms]  & 0.45(5.83) &  0.14(0.44) \\
        PR interval[ms]  & 0.51(7.57) &  0.23(0.60) \\
        QRS duration[ms]  & 0.43(5.63) &  0.20(0.56) \\
        QT interval[ms]  & 1.70(22.68) &  0.19(0.52) \\
        QT corrected interval[ms]  & 2.01(26.16) &  0.17(0.50) \\
        Rhythm ECG[bpm]  & 24.13(53.36) &  0.16(0.55) \\
        conduction ECG  & 206.93(320.21) &  0.18(0.59) \\
        Tissue alterations ECG  & 105.66(114.60) &  0.09(0.40) \\
        LVH criteria & & 0.51(0.94) \\
        \multicolumn{1}{r}{0} & 76256(93.51) & \\
        \multicolumn{1}{r}{1} & 5293(6.49) & \\
        Assessment ECG & & 0.38(0.84) \\
        \multicolumn{1}{r}{0} & 15488(25.72) & \\
        \multicolumn{1}{r}{1} & 24892(41.34) & \\
        \multicolumn{1}{r}{2} & 19839(32.94) & \\
        \bottomrule
        
    \end{tabular}
        \end{adjustbox}
    \end{minipage}%
    \hspace{0.01\textwidth}
    \begin{minipage}{.60\textwidth}
        \footnotesize
        \caption*{(b) Diagnosis codes} 
        \tiny
        \centering
        \begin{adjustbox}{width=\textwidth}
            \begin{tabular}[t]{rc}
                 \toprule        
                 Code & \# patients \\
                 \midrule
                \textbf{J41-Simple and mucopurulent chronic bronchitis} & \textbf{2507(1.58)} \\
                J41.0-Simple chronic bronchitis & 1306(0.83) \\
                J41.1-Mucopurulent chronic bronchitis & 114(0.07) \\
                J41.8-Mixed simple and mucopurulent chronic bronchitis & 100(0.06) \\
                \textbf{J42-Unspecified chronic bronchitis} & \textbf{10299(6.51)} \\
                \textbf{J43-Emphysema} & \textbf{7478(4.73)} \\
                J43.0-MacLeod syndrome & 8(0.01) \\
                J43.1-Panlobular emphysema & 16(0.01) \\
                J43.2-Centrilobular emphysema & 190(0.12) \\
                J43.8-Other emphysema & 12(0.01) \\
                J43.9-Emphysema, unspecified & 3613(2.28) \\
                \textbf{J44-Other chronic obstructive pulmonary disease} & \textbf{158204(100.00)} \\
                J44.0-Chronic obstructive pulmonary disease with acute lower respiratory infection & 574(0.36) \\
                J44.1-Chronic obstructive pulmonary disease with acute exacerbation, unspecified & 12310(7.78) \\
                J44.8-Other specified chronic obstructive pulmonary disease & 7431(4.70) \\
                J44.9-Chronic obstructive pulmonary disease, unspecified & 138890(87.79) \\
                \bottomrule
            \end{tabular}
        \end{adjustbox}
        \vspace{1em} 
        \footnotesize
        \caption*{(c) Treatments codes}
        \tiny
        \centering
        \begin{adjustbox}{width=\textwidth}
            \begin{tabular}[t]{rc}
                 \toprule        
                 Code & \# patients \\
                 \midrule
                 R03BB05-Aclidinium bromide & 11876(7.51) \\
                 R03CC12-Bambuterol & 30(0.02) \\
                 R03BA01-Beclometasone & 2136(1.35) \\
                 R03BA02-Budesonide & 47504(30.03) \\
                 R03BA08-Ciclesonide & 1358(0.86) \\
                 R03BA05-Fluticasone & 2559(1.62) \\
                 R03AC13-Formoterol & 14035(8.87) \\
                 R03AK08-Formoterol and beclometasone & 22606(14.29) \\
                 R03AK07-Formoterol and budesonide & 28729(18.16) \\
                 R03AK11-Formoterol and fluticasone & 2547(1.61) \\
                 R03BB06-Glycopyrronium bromide & 7412(4.69) \\
                 R03AC18-Indacaterol & 11115(7.03) \\
                 R03AL04-Indacaterol and glycopyrronium bromide & 12481(7.89) \\
                 R01AX03-Ipratropium bromide & 1661(1.05) \\
                 R03BB01-Ipratropium bromide & 84593(53.47) \\
                 R03AK02-Isoprenaline and other drugs for obstructive airway diseases & 2(0.00) \\
                 R03BA07-Mometasone & 726(0.46) \\
                 R03AC02-Salbutamol & 88292(55.81) \\
                 R03CC02-Salbutamol & 443(0.28) \\
                 R03AK04-Salbutamol and sodium cromoglicate & 63(0.04) \\
                 R03AC12-Salmeterol & 13633(8.62) \\
                 R03AK06-Salmeterol and fluticasone & 46172(29.19) \\
                 R03AC03-Terbutaline & 6355(4.02) \\
                 R03CC03-Terbutaline & 1454(0.92) \\
                 R03CC53-Terbutaline, combinations & 79(0.05) \\
                 R03BB04-Tiotropium bromide & 67081(42.40) \\
                 R03AK10-Vilanterol and fluticasone furoate & 3635(2.30) \\

                \bottomrule
            \end{tabular}
        \end{adjustbox}
        \vspace{1em} 
        \footnotesize
        \caption*{(c) Hospitalization codes}
        \tiny
        \centering
        \begin{adjustbox}{width=\textwidth}
            \begin{tabular}[t]{rc}
                 \toprule        
                 Code & \# patients \\
                 \midrule
                 001-139-infectious and parasitic diseases & 7164(4.53)\\
                140-239-neoplasms & 10002(6.32)\\
                 \makecell[r]{240-279-endocrine, nutritional and metabolic \\   diseases, and immunity disorders} & 21816(13.79)\\
                 \makecell[r]{280-289-diseases of the blood\\ and blood-forming organs} & 7964(5.03)\\
                290-319-mental disorders & 12871(8.14)\\
                 \makecell[r]{320-389-diseases of the nervous system \\and sense organs} &13685(8.65)\\
                390-459-diseases of the circulatory system & 25852(16.34)\\
                460-519-diseases of the respiratory system & 25555(16.15)\\
                520-579-diseases of the digestive system & 12027(7.60)\\
                580-629-diseases of the genitourinary system & 15123(9.56)\\
                 \makecell[r]{710-739-diseases of the musculoskeletal system \\ and connective tissue} &7962(5.03)\\
                 \makecell[r]{780-799-symptoms, signs, and \\ill-defined conditions} & 13991(8.84)\\
                 800-999-injury and poisoning & 7252(4.30)\\
                 800-999-injury and poisoning & 7252(4.58)\\
                \makecell[r]{V1-V90-Supplementary classification of factors\\ influencing health status}
                  & 23143(14.63)\\
                Other causes & 3415(2.16)\\
                \bottomrule
            \end{tabular}
        \end{adjustbox}
    \end{minipage} }
    \label{tab:baselines}
\end{table}

\begin{table}[ht]
    \centering
    \caption{Survivalboost, hyper-parameter search, top 10 configuration}
    \begin{tabular}{ccccc}
    \toprule
n. iterations & min samples leaf & max leaf nodes & learning rate & mean score \\
    \midrule
75      & 10                 & 40               & 0.05           & -0.1347     \\
100     & 110                & 50               & 0.05           & -0.1349     \\
100     & 30                 & 30               & 0.05           & -0.1350     \\
75      & 30                 & 50               & 0.05           & -0.1350     \\
75      & 70                 & 30               & 0.05           & -0.1350     \\
75      & 110                & 20               & 0.1            & -0.1350     \\
150     & 70                 & 20               & 0.05           & -0.1350     \\
100     & 30                 & 40               & 0.05           & -0.1351     \\
75      & 50                 & 50               & 0.05           & -0.1351     \\
200     & 10                 & 20               & 0.05           & -0.1352    \\
\bottomrule
\end{tabular}
    
    \label{tab:surv}
\end{table}

\begin{table}[ht]
    \centering
    \caption{DeepPseudo, hyper-parameter search, top 10 configuration}
    \footnotesize
\begin{tabular}{ccccccccc}
\toprule
shared-size & shared-layers & cause spec-size & cause spec-layers & dropout & batch size & loss1 & loss2 & loss \\
\midrule
200 & 1 & 100 & 3 & 0   & 128  & 0.78  & 5.053 & 5.8330\\
200 & 1 & 100 & 3 & 0   & 128  & 0.78  & 5.053 & 5.8331\\
100 & 3 & 50  & 4 & 0   & 256  & 0.78  & 5.053 & 5.8332\\
100 & 3 & 50  & 4 & 0   & 256  & 0.781 & 5.053 & 5.8332\\
100 & 3 & 50  & 4 & 0   & 256  & 0.78  & 5.054 & 5.8338\\
100 & 1 & 50  & 3 & 0   & 128  & 0.781 & 5.053 & 5.8344\\
300 & 4 & 200 & 3 & 0   & 128  & 0.781 & 5.054 & 5.8349\\
100 & 3 & 50  & 4 & 0   & 256  & 0.781 & 5.054 & 5.8350\\
300 & 4 & 200 & 3 & 0   & 128  & 0.782 & 5.053 & 5.8352\\
300 & 2 & 100 & 3 & 0.1 & 128  & 0.781 & 5.054 & 5.8356\\
\midrule
\end{tabular}
    
    \label{tab:deeppseudo}
\end{table}

\begin{table}[ht]
    \centering
    \caption{SurvTRACE, hyper-parameter search, top 10 configuration}
    \begin{tabular}{ccccc}
    \toprule
n. layers & embedding size & inter. size & n. att. heads & mean score \\
    \midrule
2      & 96               & 64             & 12            & 1.654     \\
2      & 24               & 64             & 12            & 1.661     \\
2      & 48               & 64             & 6             & 1.746     \\
1      & 12               & 64             & 6             & 1.779     \\
2      & 12               & 64             & 6             & 1.784     \\
2      & 96               & 16             & 3             & 2.741     \\
2      & 96               & 64             & 6             & 5.239     \\
1      & 24               & 64             & 6             & 5.280     \\
3      & 48               & 64             & 6             & 5.486     \\
\bottomrule
\end{tabular}
    
    \label{tab:survtrace}
\end{table}

\begin{table}[ht]
    \centering
    \caption{DDH, hyper-parameter search, top 10 configuration}
    \footnotesize
\begin{tabular}{ccccccccccc}

\toprule
n. rnn& type& size rnn&dropout&n. attn.&cause spec. & n. cause spec.&  $\alpha$&$\beta$&$\sigma$ \\

\midrule
2 & GRU  & 100 & 0.1 & 350 & 300 & 5 & 1   & 0.1 & 3   & 192.869 \\
2 & GRU  & 350 & 0.3 & 200 & 200 & 2 & 0.1 & 1   & 1   & 193.223 \\
1 & LSTM & 100 & 0.1 & 100 & 350 & 3 & 0.1 & 0.1 & 3   & 193.227 \\
2 & GRU  & 100 & 0.3 & 200 & 350 & 4 & 0.1 & 0.1 & 3   & 193.356 \\
2 & LSTM & 100 & 0.1 & 200 & 200 & 4 & 0.1 & 1   & 1   & 193.853 \\
3 & GRU  & 350 & 0.1 & 200 & 300 & 5 & 0.1 & 5   & 0.1 & 193.977 \\
1 & GRU  & 350 & 0.3 & 300 & 300 & 3 & 0.1 & 3   & 0.1 & 193.993 \\
1 & LSTM & 200 & 0.1 & 350 & 350 & 3 & 3   & 5   & 5   & 194.1   \\
3 & GRU  & 200 & 0.1 & 100 & 300 & 4 & 0.1 & 1   & 3   & 194.12  \\
1 & GRU  & 300 & 0.6 & 300 & 350 & 4 & 3   & 1   & 3   & 194.345 \\
\bottomrule
\end{tabular}

    \label{tab:ddh}
\end{table}

\begin{table}[ht]
    \centering
    \caption{DRSM, hyper-parameter search, top 10 configuration}
    \footnotesize
\begin{tabular}{cccccccc}
\toprule
distribution & k & type rnn & hidden & n. layers rnn & drop & drop rnn & avg loss \\
\midrule
LogNormal    & 5 & GRU  & 300        & 2            & 0    &  0.1     & 1.316735 \\
LogNormal    & 5 & GRU  & 400        & 3            & 0    & 0        & 1.316914 \\
LogNormal    & 3 & GRU  & 400        & 2            & 0    &  0.3     & 1.316973 \\
LogNormal    & 3 & GRU  & 350        & 2            & 0.1  &  0.1     & 1.317143 \\
LogNormal    & 4 & GRU  & 200        & 2            & 0    & 0        & 1.317208 \\
LogNormal    & 5 & GRU  & 300        & 2            & 0.3  & 0        & 1.31725  \\
LogNormal    & 3 & GRU  & 300        & 2            &  0.1    &  0.1        & 1.317261 \\
LogNormal    & 3 & GRU  & 350        & 3            & 0    & 0        & 1.317278 \\
LogNormal    & 5 & GRU  & 400        & 1            &  0.1    & 0        & 1.317281 \\
LogNormal    & 3 & GRU  & 350        & 3            &  0.1    & 0        & 1.317305 \\
\bottomrule
\end{tabular}

    \label{tab:drsm}
\end{table}

\end{document}